\documentclass[runningheads]{llncs}
\usepackage[T1]{fontenc}
\usepackage{graphicx}
\usepackage{makecell}
\usepackage{multirow}
\usepackage{multicol}
\usepackage{booktabs}
\usepackage{url}
\usepackage{amsmath,amssymb,amsfonts}

\usepackage{xcolor}  
\definecolor{myRed}{RGB}{195,10,10}
\definecolor{myGreen}{RGB}{55,149,73}
\definecolor{myBlue}{RGB}{5,5,60}

\usepackage[misc]{ifsym}

\begin{document}
\title{
Temporally-Constrained Video Reasoning Segmentation and Automated Benchmark Construction
}
\titlerunning{Temporally-Constrained Video Reasoning Segmentation}
%

\author{Yiqing Shen \and Chenjia Li \and Chenxiao Fan \and Mathias Unberath\textsuperscript{(\Letter)}}

\institute{Johns Hopkins University, Baltimore, MD, USA\\
\email{\{yshen92,unberath\}@jhu.edu}}

\authorrunning{Y. Shen et al.}
%
%
\maketitle              

\begin{abstract}

Conventional approaches to video segmentation are confined to predefined object categories and cannot identify out-of-vocabulary objects, let alone objects that are not identified explicitly but only referred to implicitly in complex text queries. This shortcoming limits the utility for video segmentation in complex and variable scenarios, where a closed set of object categories is difficult to define and where users may not know the exact object category that will appear in the video. Such scenarios can arise in operating room video analysis, where different health systems may use different workflows and instrumentation, requiring flexible solutions for video analysis.
Reasoning segmentation (RS) now offers promise towards such a solution, enabling natural language text queries as interaction for identifying object to segment. However, existing video RS formulation assume that target objects remain contextually relevant throughout entire video sequences.
This assumption is inadequate for real-world scenarios in which objects of interest appear, disappear or change relevance dynamically based on temporal context, such as surgical instruments that become relevant only during specific procedural phases or anatomical structures that gain importance at particular moments during surgery.
To enable more research on RS for dynamic tasks, our first contribution is the introduction of \textbf{temporally-constrained video reasoning segmentation}, a novel task formulation that requires models to implicitly infer when target objects become contextually relevant based on text queries that incorporate temporal reasoning.
However, we do not know how well method perform this task, because we do not have a dataset to study this. So the first step is to construct a dataset.
Since manual annotation of temporally-constrained video RS datasets would be expensive and limit scalability, our second contribution is an innovative automated benchmark construction method.
Finally, we present \textit{TCVideoRSBenchmark}, a temporally-constrained video RS dataset containing 52 samples using the videos from the MVOR dataset.
The TCVideoRSBenchmark is available at \url{https://github.com/arcadelab/TCVideoRSBenchmark}.

\keywords{Video Analysis \and Reasoning Segmentation \and Digital Twin Representation \and Large Language Model (LLM) Agent \and Benchmark.}
\end{abstract}

\section{Introduction}

Conventional video segmentation task formulations, including semantic segmentation and instance segmentation, are fundamentally limited by their confinement to predefined object categories and their inability to respond to text queries that require understanding of implicit relationships and multi-step reasoning for object identification \cite{grammatikopoulou2024spatio}. 
These limitations restrict their applicability in dynamic clinical environments, such as operating room (OR) video analysis for monitoring surgical workflow, which requires the ability to respond to context-dependent queries that go beyond simple object identification, encompassing complex procedural understanding that traditional segmentation methods cannot provide.
Reasoning segmentation (RS) \cite{lisa} enables text-based object identification and has shown promise to enhance user interaction in surgical workflow analysis \cite{shen2025operating,shen2025reasoning}. 
However, existing video RS methods operate under a critical assumption that target objects remain contextually relevant throughout entire video sequences. 
This assumption becomes inadequate for real-world applications where objects of interest appear, disappear, or change relevance dynamically based on temporal context. 
In other words, current video RS approaches cannot effectively handle queries such as ``\textit{segment the anesthesia equipment only during the patient preparation phase}'' that require understanding of temporal boundaries \cite{shen2025reasoning}.

This temporal limitation undermines the potential of RS for applications that require precise temporal understanding, where these video monitoring frameworks must understand not only what and where objects are located, but also precisely when they become relevant within specific procedural contexts \cite{jin2021temporal,xu2024sedmamba}.
In surgical workflows, for instance, procedures exhibit inherently structured temporal organization with distinct phases such as patient preparation, anesthesia induction, surgical intervention, and recovery, each characterized by different sets of relevant objects and personnel configurations \cite{demir2024towards}. 
The importance of temporal relationships extends beyond simple phase identification to encompass complex dependencies between procedural events and object relevance periods, where the same instrument or personnel may require different analytical attention depending on the current procedural context.
Despite this need, the temporal dimension remains largely unexplored in current RS literature due to the absence of appropriate benchmarks.

\begin{figure}[t!]
\centering
\includegraphics[width=0.9\linewidth]{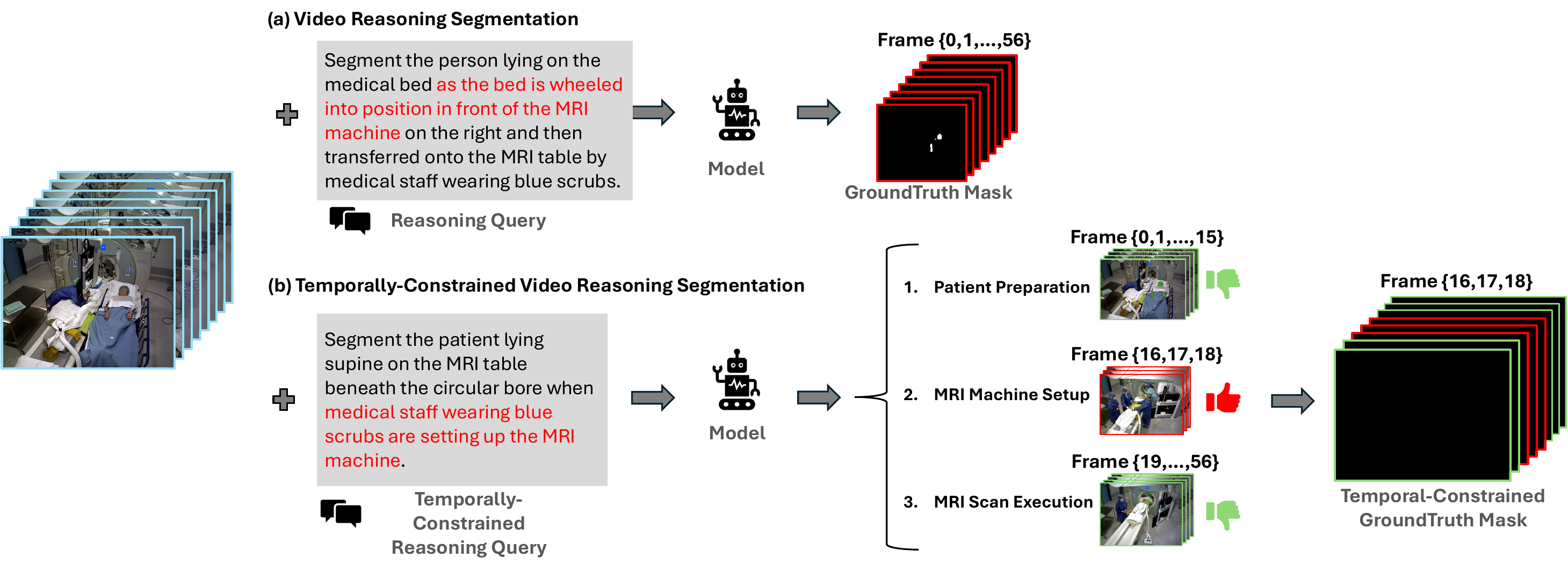}
\caption{
Comparison between conventional video RS and the proposed temporally-constrained video RS task formulation. 
(a) Conventional video RS processes implicit text queries across entire video sequences, generating segmentation masks for all frames regardless of temporal relevance. 
(b) Temporally-constrained video reasoning segmentation restricts segmentation to specific temporal boundaries. 
The example demonstrates segmenting a patient only during the ``MRI Machine Setup'' phase (frames 16-18) rather than throughout the entire video sequence. 
}\label{fig:intro}
\end{figure}

To address this gap, we first introduce a novel task formulation termed \textbf{temporally-constrained video reasoning segmentation}, as illustrated in Fig.~\ref{fig:intro}.
This task formulation extends video RS beyond continuous object tracking by incorporating phase-specific or action-specific temporal constraints to perform segmentation.
For this new task, due to the lack of appropriate dataset, we do not know how model performs. 
Consequently, the initial step is to construct a benchmark dataset.
Correspondingly, we propose an automated benchmark construction method that leverages digital twin (DT) representations, defined as structured intermediate representations that preserve semantic, spatial, and temporal relationships between entities and their interactions \cite{shen2025position}, combined with large language models (LLMs) to generate temporally aware implicit queries without requiring manual annotation efforts that would otherwise limit the scalability of the dataset. 
Unlike previous applications of digital twin representations that primarily utilized semantic and spatial information for general reasoning tasks \cite{shen2025rvtbench}, our approach specifically exploits the temporal dimension embedded within DT structures to construct queries that require understanding of when objects become relevant within surgical workflow phases, enabling the generation of temporally-constrained reasoning queries that reflect the dynamic nature of OR procedures.

The major contributions are three-fold.
First, we propose the temporally-constrained video RS, which is a new task that requires models to perform RS only within specified temporal boundaries. 
Second, we develop an automated pipeline that constructs benchmark datasets through DT representations and LLM-based query generation, enabling the scalable creation of temporally-constrained reasoning queries. 
Third, we construct a benchmark dataset for temporally-constrained video RS (namely \textit{TCVideoRSBenchmark}), which contains 52 samples that span various surgical scenarios and temporal reasoning.

\section{Task Formulation}

We formalize temporally-constrained video RS as an extension of traditional video RS that incorporates temporal boundaries derived from surgical workflow phases. 
Unlike conventional video RS that assumes the presence and relevance of continuous objects throughout entire video sequences, our formulation acknowledges that surgical procedures exhibit a structured temporal organization where objects, personnel, and equipment become contextually relevant only during specific procedural phases.
Specifically, given an input video sequence $\mathcal{V} = \{I^{(1)}, I^{(2)}, \ldots, I^{(T)}\}$ consisting of frames $T$ and an implicit reasoning query $Q$ that describes the target segmentation objective, traditional video reasoning segmentation seeks to produce a sequence of binary segmentation masks $\mathcal{M} = \{M^{(1)}, M^{(2)}, \ldots, M^{(T)}\}$ where each $M^{(t)} \in \{0,1\}^{H \times W}$ indicates the object in pixels at the timestep $t$. 
However, this formulation assumes that the reasoning query $Q$ remains equally applicable across all temporal instances, which proves inadequate for surgical workflow analysis where contextual relevance varies across procedural phases.

In contrast, temporally-constrained video RS requires models to implicitly infer temporal boundaries from the reasoning query itself, determining when the segmentation objective becomes contextually applicable without explicit temporal annotations.
Therefore, the model must parse the implicit query $Q$ to extract both the target segmentation objective and the underlying temporal constraints embedded within the query semantics.
Formally, we define a temporal constraint $\tau_Q: \mathbb{N} \rightarrow \{0,1\}$ that the model must learn to determine the validity of the reasoning query at each time step based solely on the query content and the video context:
\begin{equation}
\tau_Q(t) = \begin{cases} 
1 & \text{if } f_{\text{inference}}(Q, \mathcal{V}, t) = \text{active} \\
0 & \text{otherwise}
\end{cases}
\end{equation}
where $f_{\text{inference}}$ represents the temporal reasoning capacity of the model that analyzes the query $Q$, the video sequence $\mathcal{V}$, and the current time step $t$ to determine whether segmentation should be performed. 
The temporally-constrained reasoning segmentation task then becomes:
\begin{equation}
\mathcal{M}_{\text{constrained}} = \{M^{(t)} \cdot \tau_Q(t) | t = 1, 2, \ldots, T\}.
\end{equation}
This definition ensures that segmentation masks are produced only during periods that the model infers as temporally relevant from the query, with $M^{(t)} = \emptyset$ (empty mask) when $\tau_Q(t) = 0$.

\section{Dataset Construction}

\begin{figure}[t!]
\centering
\includegraphics[width=\linewidth]{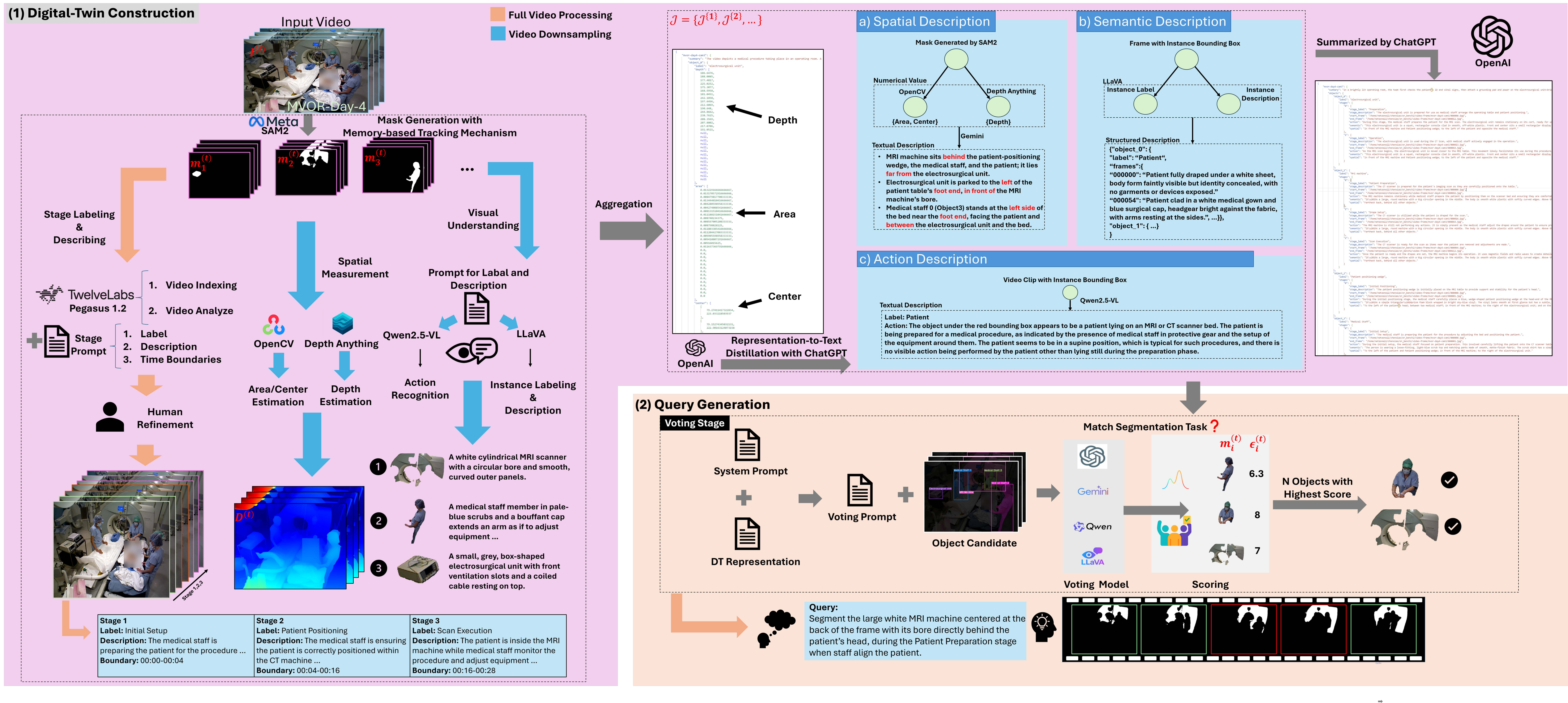}
\caption{
Overview of the proposed automated pipeline for temporally-constrained video RS benchmark construction. 
The framework consists of two primary components: (1) Digital twin construction, which transforms raw operating room video sequences into structured representations through specialized vision foundation models, including TwelveLabs Pegasus 1.2 for action and phase identification, SAM2 \cite{sam2} for instance segmentation, DepthAnything2 \cite{depthanything} for spatial measurement, and LLaVA \cite{llava} for semantic understanding; and (2) Query generation, which employs an ensemble voting by LLM to identify salient object candidates and their associated procedural phases from the digital twin representation, followed by template-based synthesis of temporally-constrained reasoning queries that embed implicit temporal boundaries within natural language formulations. 
}\label{fig:method}
\end{figure}

\subsubsection{Digital Twin Representation for Temporal Reasoning}
Although prior work on DT representations \cite{shen2025rvtbench} for automatic reasoning data generation focuses on general visual reasoning tasks, our approach introduces surgical workflow-aware temporal modeling that explicitly captures temporal information such as procedural stage transitions and temporal constraints. 
Our dataset construction begins with the transformation of the OR video sequences into structured DT representations that not only preserve semantic and spatial relationships but also encode temporal boundaries for analysis of surgical workflow.
Formally, given an OR video sequence $\mathcal{V} = \{I^{(1)}, I^{(2)}, \ldots, I^{(T)}\}$, we construct a corresponding DT representation $\mathcal{J} = \{\mathcal{J}^{(1)}, \mathcal{J}^{(2)}, \ldots, \mathcal{J}^{(T)}\}$ where each frame-level representation $\mathcal{J}^{(t)}$ encodes multi-dimensional information through a suite of vision models $\Omega = \{\omega_1, \omega_2, \ldots, \omega_K\}$, which extract information, formally expressed as $\mathcal{J}^{(t)} = \Omega(I^{(t)})$.
Temporal information extraction begins with the identification of the surgical stage at the video level through TwelveLabs Pegasus 1.2\footnote{\url{https://www.twelvelabs.io/blog/introducing-pegasus-1-2}}, a multimodal foundation model for video indexing and analysis to decompose input video sequences into distinct procedural phases $\Phi = \{\phi_1, \phi_2, \ldots, \phi_P\}$ in the full frame video stream. 
For each phase, TwelveLabs Pegasus 1.2 extracts three components, namely descriptive labels that categorize procedural activity, detailed descriptions that capture specific actions and interactions between objects that occur during the stage, and the corresponding temporal boundaries that define the temporal extent of each phase. 
Subsequently, this automated extraction is refined through human verification to ensure accuracy and clinical relevance.
For semantic and spatial encoding, we employ SAM2 \cite{sam2} for example identification and segmentation, generating masks $\mathcal{M}^{(t)} = \{m_i^{(t)}\}_{i=1}^{N^{(t)}}$ ($t=1,\cdots,T$) where each $m_i^{(t)}$ represents a binary mask for object $i$ with confidence score $\epsilon_i^{(t)}$ at time step $t$. 
The notation $N^{(t)}$ denotes the number of objects detected in frame $t$, and $\epsilon_i^{(t)} \in [0,1]$ quantifies the confidence in the segmentation of object $i$ derived from SAM2 \cite{sam2}. 
For temporal coherence across frames, we leverage SAM2's memory-based tracking mechanism:
\begin{equation}
m_i^{(t+k)} = \text{SAM}_\text{track}(I^{(t+k)}, \{m_i^{(t+k')}\}_{k'=0}^{k}), \quad 0 < k < t_s
\end{equation}
where $I^{(t+k)}$ represents the target frame for tracking, $\{m_i^{(t+k')}\}_{k'=0}^{k}$ denotes the sequence of previous masks used for temporal propagation, and $t_s$ represents the temporal sampling interval for key frame processing to balance computational efficiency with tracking accuracy.
Additional spatial information is complemented $\mathcal{M}^{(t)}$ through DepthAnything2 \cite{depthanything}, which generates dense depth maps $D^{(t)} \in \mathbb{R}^{H \times W}$ for each frame of resolution $H \times W$. 
For every object $i$, we can compute the depth statistics within its mask region as $d_i^{(t)} = \{D^{(t)}(p) | p \in m_i^{(t)}\}$, where $p$ represents the coordinates of the pixels within the mask. 
The mean depth $\mu_i^{(t)} = \frac{1}{|m_i^{(t)}|} \sum_{p \in m_i^{(t)}} D^{(t)}(p)$ and the standard deviation $\sigma_i^{(t)} = \sqrt{\frac{1}{|m_i^{(t)}|} \sum_{p \in m_i^{(t)}} (D^{(t)}(p) - \mu_i^{(t)})^2}$ characterize the spatial positioning and the depth variation, where $|m_i^{(t)}|$ denotes the number of pixels in the mask.
Finally, we also employ OpenCV operators to extract complementary visual features, including optical flow vectors for motion analysis, color histograms for appearance characterization, and texture descriptors following previous work \cite{shen2025rvtbench}.
We also include a description of temporal action at the object level to encode temporal information at the object level as a complement to semantic understanding and temporal information at the video level, where we generate stage-specific action sequences $\mathcal{A}^{(t)} = \{a_{i,j}^{(t)}\}_{j=1}^{M_i^{(t)}}$ for each object $i$ via Qwen2.5-VL \cite{bai2025qwen2}. 
These action descriptions explicitly encode when specific activities begin, progress, and conclude, enabling the DT representation to understand temporal relevance of object within procedural contexts. 
Semantic understanding is achieved through LLaVA \cite{llava}, which generates object-level descriptors $\mathcal{S}^{(t)} = \{s_i^{(t)}\}_{i=1}^{N^{(t)}}$ that capture object attributes, functional roles, and contextual relationships within the surgical environment. 

\begin{figure}[t!]
\centering
\includegraphics[width=\linewidth]{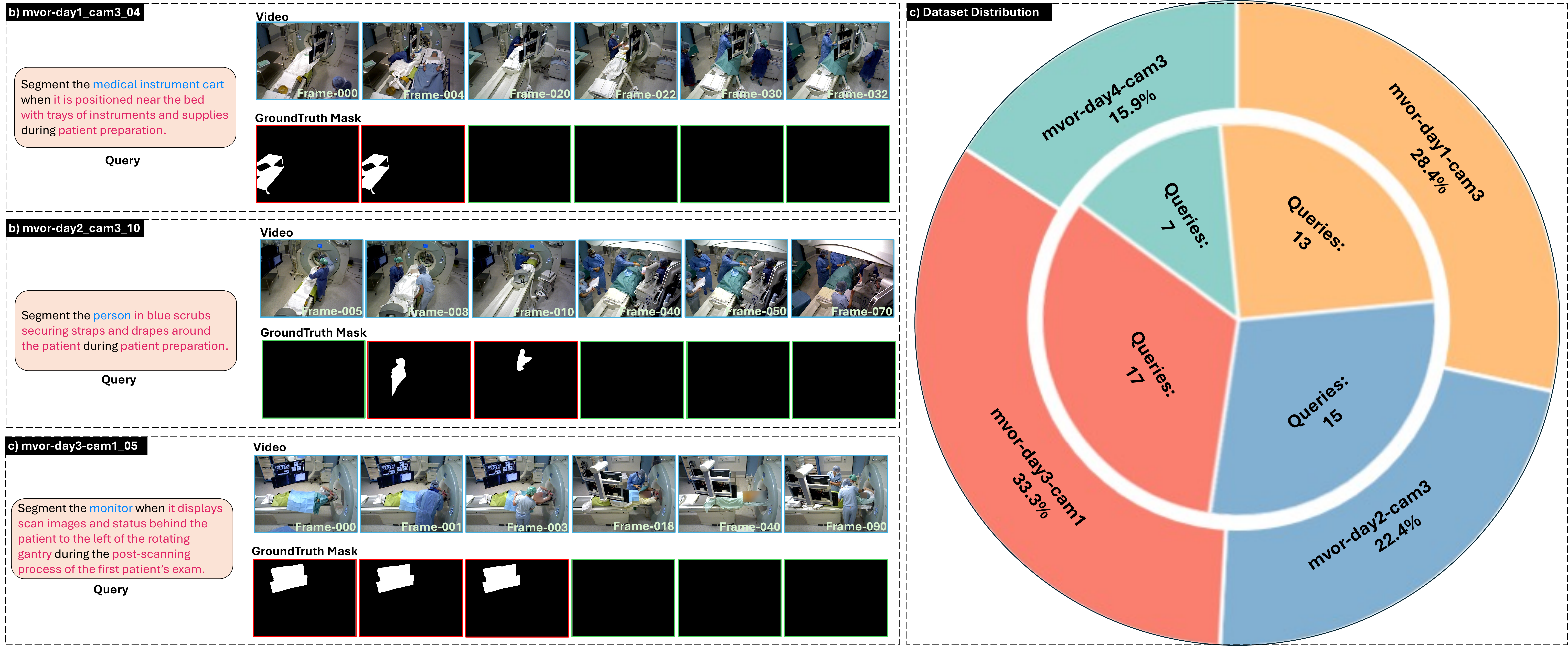}
\caption{
Overview of \textit{TCVideoRSBenchmark} dataset composition and representative examples of temporally-constrained video RS. 
(a-c) Three exemplar queries demonstrating temporal constraint reasoning across different procedural phases: (a) segmenting medical instrument cart positioned near the bed during patient preparation (mvor-day1\_cam3\_04), (b) segmenting personnel in blue scrubs securing patient drapes during patient preparation (mvor-day2\_cam3\_10), and (c) segmenting monitor displaying scan images during post-scanning process (mvor-day3-cam1\_05). 
Each example includes the temporally-constrained reasoning query, corresponding video frame sequence, and ground truth segmentation masks that are active only during the specified temporal boundaries. 
(d) Dataset distribution across four MVOR videos, totaling 52 temporally-constrained video RS samples.
}\label{fig:data}
\end{figure}

\subsubsection{Query Generation with LLM}

Given the constructed DT representation $\mathcal{J} = \{\mathcal{J}^{(1)}, \mathcal{J}^{(2)}, \ldots, \mathcal{J}^{(T)}\}$ per video, we then employ an LLM-based agent framework to generate the temporally-constrained RS queries that embed implicit temporal boundaries using $\mathcal{J}$. 
Specifically, it consists of three sequential stages: (1) identification of the object candidate through ensemble voting, (2) verification of temporal alignment, and (3) generation of queries.
The first stage leverages an ensemble voting based on multiple LLMs to identify potential objects and their associated temporal phases from the DT representation for subsequent query generation.
Specifically, we deploy LLM instances with different prompts $\{\mathcal{L}_1, \mathcal{L}_2, \ldots, \mathcal{L}_K\}$ to independently evaluate object candidates and their relevance for temporally-constrained video RS task. 
For each object $i$ identified in the DT representation at time step $t$, we define a voting function by LLM:
\begin{equation}
v_k^{(i,t)} = \mathcal{L}_k(\mathcal{J}^{(t)}, s_i^{(t)}, \phi_t),
\end{equation}
where $s_i^{(t)}$ represents the object-level semantic description of object $i$ at time $t$, and $\phi_t$ denotes the video-level phase in the corresponding temporal window. 
Each LLM $\mathcal{L}_k$ assigns a relevance score $v_k^{(i,t)} \in [0, 1]$ based on the semantic attributes, spatial configuration, and temporal context of the object within the identified surgical phase.
The ensemble voting result for each candidate object-phase pair $(i, \phi)$ is computed through weighted aggregation $V^{(i,\phi)} = \frac{1}{K} \sum_{k=1}^{K} \sum_{t \in T_\phi} v_k^{(i,t)} \cdot w_k$, where $T_\phi$ represents the temporal window corresponding to phase $\phi$, and $w_k$ denotes the weight assigned to LLM $\mathcal{L}_k$ (where we set $w_k=\frac{1}{K}$ in this work for simplicity). 
All objects with voting scores exceeding a pre-determined threshold $\theta_\text{vote}$ are selected as candidates for subsequent query generation.
Therefore, we can have multiple objects selected from this stage.
The second stage performs temporal alignment verification to ensure coherence between selected objects and their associated temporal phases. 
For each pair of candidates $(i^*, \phi^*)$ selected from the voting process, we verify temporal consistency through:
\begin{equation}
\tau_{align}(i^*, \phi^*) = \mathbf{1}\left[\exists t \in T_{\phi^*} : \epsilon_i^{(t)} > \theta_{conf} \wedge a_i^{(t)} \neq \emptyset\right],
\end{equation}
where $\epsilon_i^{(t)}$ represents the confidence score of the object $i$ at time $t$, $a_i^{(t)}$ denotes the description of the temporal action of the object and $\theta_{conf}$ is the confidence threshold. 
It ensures that selected objects exhibit a meaningful presence and activity during their associated procedural phases.
The final stage synthesizes temporally-constrained video RS queries by embedding implicit temporal boundaries within natural language formulations. 
Given a validated object-phase pair $(i^*, \phi^*)$, we construct the reasoning query through template-based generation:
\begin{equation}
Q_{temp} = \mathcal{G}(s_{i^*}, \phi^*, T_{\phi^*}, \mathcal{R}_{spatial}, \mathcal{R}_{semantic}),
\end{equation}
where $\mathcal{G}$ represents the LLM for query generation, $\mathcal{R}_{spatial}$ and $\mathcal{R}_{semantic}$ denote the spatial and semantic relationship descriptors extracted from the DT representation. 
The corresponding ground truth RS masks are extracted from the DT representation by applying temporal constraints to the pre-existing instance masks. 
Specifically, for the selected object $i^*$ and phase $\phi^*$, the temporally-constrained ground truth is constructed as $\mathcal{M}_{gt} = \{M^{(t)}_{i^*} \cdot \tau_{\phi^*}(t) | t = 1, 2, \ldots, T\}$, 
where $M^{(t)}_{i^*}$ represents the instance mask for object $i^*$ at time $t$ stored in the DT representation, and $\tau_{\phi^*}(t)$ is the temporal constraint function that equals 1 when $t \in T_{\phi^*}$ and 0 otherwise. 
Finally, all the generated samples (the triplet query, video, and masks) go through manual verification, and we filter the incorrect or improper ones. 

\subsubsection{Dataset Statistics}

The video sequences in TCVideoRSBenchmark are from the MVOR dataset \cite{srivastav2018mvor}, which is an authentic OR dataset, consisting of 732 synchronized frames at $640\times480$ resolution. 
TCVideoRSBenchmark comprises 52 temporally-constrained video RS samples from 4 representative videos, with the following distribution mvor-day1\_cam3 contributes 15 queries (28.4\%), mvor-day2\_cam3 provides 13 queries (22.4\%), mvor-day3\_cam1 contains 17 queries (33.3\%), and mvor-day4\_cam3 includes 7 queries (15.9\%), as illustrated in Fig.~\ref{fig:data}.
Each sample in \textit{TCVideoRSBenchmark} consists of a temporally-constrained reasoning query paired with video and corresponding ground truth segmentation masks that are temporally bounded to specific procedural phases. 
The queries encompass diverse surgical workflow scenarios, where representative examples in Fig.~\ref{fig:data} include segmenting medical instrument carts during patient preparation phases.
The temporal constraints embedded within the queries span multiple procedural phases commonly observed in operating room workflows, including patient preparation, equipment setup, active intervention periods, and post-procedural activities. 

\section{Conclusion}

The innovative formulation of temporally-constrained video RS aims to address the conventional RS's limitation of assuming continuous object tracking. 
As a new task, we propose an automated benchmark construction pipeline that uses DT representations, demonstrating the potential to create scalable benchmark datasets for temporally-constrained video RS without manual annotation efforts. 
Based on this method, the \textit{TCVideoRSBenchmark}, including 52 samples derived from the MVOR dataset, is constructed to allow evaluation of the ability to understand when segmentation objectives become contextually applicable within surgical procedures.
Overall, the demonstration of temporally-constrained video RS opens new possibilities for surgical workflow monitoring methods that provide contextually relevant insights aligned with the natural temporal structure of medical procedures.

\subsection*{Acknowledgments}
This work was supported in part by the JHU Amazon Initiative for Artificial Intelligence (AI2AI) fellowship program and NSF CAREER award (NSF Award No. 2239077). 

\bibliographystyle{plain}
\bibliography{main}

\end{document}